\documentclass{article}

\usepackage{bm}
\usepackage{ascmac}
\usepackage{algorithm}
\usepackage{algorithmic}

\usepackage{PRIMEarxiv}

\usepackage[utf8]{inputenc} 
\usepackage[T1]{fontenc}    
\usepackage{hyperref}       
\usepackage{url}            
\usepackage{booktabs}       
\usepackage{amsfonts}       
\usepackage{nicefrac}       
\usepackage{microtype}      
\usepackage{lipsum}
\usepackage{fancyhdr}       
\usepackage{graphicx}       
\graphicspath{{media/}}     

\title{Improved Solution Search Performance of Constrained MOEA/D Hybridizing Directional Mating and Local Mating
\thanks{Presented at ISMSI2023}
}

\author{
  Masahiro Kanazaki, Takeharu Toyoda \\
  Department of Aeronautics and Astronautics, Graduate School of System Design  \\
  Tokyo Metropolitan University \\
  Tokyo\\
  \texttt{kana@tmu.ac.jp} \\
}

\begin{document}
\maketitle

\begin{abstract}
In this study, we propose an improvement to the direct mating method, a constraint handling approach for multi-objective evolutionary algorithms, by hybridizing it with local mating. Local mating selects another parent from the feasible solution space around the initially selected parent.
The direct mating method selects the other parent along the optimal direction in the objective space after the first parent is selected, even if it is infeasible. It shows better exploration performance for constraint optimization problems with coupling NSGA-II, but requires several individuals along the optimal direction. Due to the lack of better solutions dominated by the optimal direction from the first parent, direct mating becomes difficult as the generation proceeds.
To address this issue, we propose a hybrid method that uses local mating to select another parent from the neighborhood of the first selected parent, maintaining diversity around good solutions and helping the direct mating process. We evaluate the proposed method on three mathematical problems with unique Pareto fronts and two real-world applications. We use the generation histories of the averages and standard deviations of the hypervolumes as the performance evaluation criteria. 
Our investigation results show that the proposed method can solve constraint multi-objective problems better than existing methods while maintaining high diversity.
\end{abstract}


\section{Introduction}
\label{sec:intro}
Real-world design problems typically involve multiple constraints, and various constraint-handling schemes have been proposed for evolutionary algorithms (EAs). 
Some methods generate solutions repeatedly until a feasible solution is found and retained\cite{hoffmeister1996problem}, while others use penalty functions to quantify the degree of constraint violation\cite{michalewicz1996evolutionary}. However, these approaches often eliminate solutions that violate constraints, even if they have excellent objective values and could contribute to better solutions.

\begin{figure}
    \centering
    \includegraphics[scale=1.0] {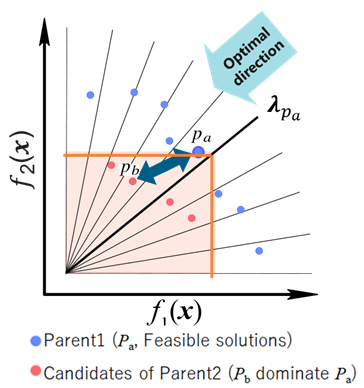}
    \caption{Schematic illustration of DM. The constraint-violating solution in the red circle with a better objective function value than that of CMOEA/D is a candidate for the parent solution.}
    \label{fig:illustDM}
\end{figure}

To address this issue, directed mating (DM) \cite{miyakawa2013two} has been proposed as a way to maintain constraint-violating solutions that are superior in the objective space. 
DM selects a second parent based on the optimal direction in the objective space, even if it violates constraints. Figure \ref{fig:illustDM} illustrates the DM process. 
DM has been integrated into MOEA/D as CMOEA/D-DM with Archives (CMOEA/D-DMA)\cite{miyakawa2018directed} which archives superior solutions in the objective space, even if they violate constraints. 
However, CMOEA/D-DMA requires a sufficient number of better solutions to maintain DM's exploration performance, and this can be challenging for real-world problems with high evaluation costs.

To improve CMOEA/D-DMA's performance for real-world applications, we introduce neighborhood mating (LM) to cultivate better solutions, whether they are feasible or not. 
Specifically, we hybridize LM with DM to improve DM's operation, especially at the end of a generation when better solutions are scarce. 
In this paper, we describe our approach in detail and present experimental results for mathematical problems and real-world applications. We organize the paper as follows: Section \ref{sec:intro} provides background and motivation for our study, Section \ref{sec:Reviewconst} describes the constraint handling techniques used in MOEA/D and the DM algorithm, Section 3 presents our proposed improvement to DM using LM, Section 4 describes our numerical experimental setup, Section 5 presents the numerical results and discussions, and Section 6 concludes our study.

\section{CONSTRAINED OPTIMIZATIONS BY EVOLUTIONARY COMPUTATION}
\label{sec:Reviewconst}
\subsection{Constrained Multi-Objective Evolutionary Algorithm}

The evolutionary algorithm (EA)\cite{van1998multiobjective} is a meta-heuristic optimization algorithm inspired by life's evolution, and the multi-objective EA is enhanced by Pareto optimality. 
It can solve the multi-objective optimization problem as expressed as following equation by Pareto ranking or scalarizing the design problem. 

\begin{displaymath}
\left\{
\begin{array}{l}
{\rm minimize} \ \ f_j({\bm x}) \\
 \ \ {\rm Subject \ to} \ \ h_k({\bm x}) < 0 \\
j=1,...m, \ k=1, ...,p
\end{array}
\right.
\end{displaymath}
    
\noindent
Where ${\bm x}$ is a vector of design variable, $f_j({\bm x})$ is a $j$th objective function, and $h_k({\bm x})$ is a $k$th constraint violation. $m$ is the number of objective functions and $p$ is the number of constraint.
It has the advantage of simultaneous multi-point search, which enables a global search for a Pareto solution set for a multi-objective problem and does not require differentiability of the objective function and constraints. 
The MOEA starts with initialization to generate initial individuals in the given design space. 
Then the objective function values are calculated based on the given design variables to evaluate the individuals. 
This evaluation value is used as an objective function to determine the superiority or inferiority of a solution. 
The next-generation individuals are then created through crossover and mutation after the selection of the parents. 
This process is repeated until the termination condition is satisfied.
\par
A MOEA algorithm which is widliy used is multi-objective evolutionary algorithm with Decomposition (MOEA/D) \cite{zhang2007moea}. 
MOEA/D separates the vector of objective functions by $m$ weight vectors ${\bm \lambda}=(\lambda _1 ...\lambda_m)$ and defines the multi-objective optimization problem to the single-objective optimization problem for each weight vector ${\bm \lambda}$. 
The weighted sum function\cite{gandibleux2006multiple} and the weighted Tchebycheff function \cite{zhang2007moea} as shown in Eq. \ref{eq:tcheb} are widely used as scalarized functions. 

\begin{equation}
\label{eq:tcheb}
    g({\bm x}|{\bm \lambda})=\max_{0 \leq j \leq 1} \{\lambda^j |f_j({\bm x})-x_j\} 
\end{equation}

\noindent
$z_j,(j=1,2,..,m)$ is the reference point used in MOEA/D and is the best value of the objective function of the solution obtained during the optimization calculation. Offsprings are generated by parents pair selected from the neighbourhood of the focused weight vector, $\mathcal{B}(i)$. $i$ is a focused index.
The features of MOEA/D include the following: the diversity of solutions can be maintained by the weight vector, the search direction of the solution can be biased by changing the weight vectors, and good exploration performance even in many objective optimizations with four or more objectives compared to the Non-Dominated Sorting Genetic Algorithm-II (NSGA-II)\cite{deb2002fast}. 
\par
Constraint handling was implemented to CMOEA/D and it was proposed as constraint Multi-objective Evolutionary Algorithm with Decomposition (CMOEA/D) \cite{jain2013evolutionary}. 
Figure \ref{fig:flowCMOEAD} shows the flow chart of CMOEA/D and 
Algorithm 1 shows the solution update by CMOEA/D. 
Based on Algorithm 1, feasible solutions are always superior to infeasible solutions, even if there can be infeasible solutions that are good in the objective space on the boundary between the feasible and the infeasible. This method eliminates solutions based on the sum $\Omega({\bm x})$ of the constraint violation values before calculating the scalarized objective function values based on the weight vector.

\subsection{Implementation of Direct Mating with Archive}
When evaluating solutions obtained during optimization, CMOEA/D, or Constraint NSGA-II\cite{deb2002fast}, first evaluates solutions by the amount of constraint violation; then, the objective functions are compared if the degree of violation is the same. In these methods, infeasible solutions are evaluated as inferior to feasible solutions regardless of the superiority of the objective function value. Therefore, the infeasible solution information cannot be utilized in the optimization process, even if they have design variables that can improve the objective function near the boundary of the feasible and infeasible regions. This suggests that constraints are severe and feasible solutions are difficult to obtain. It is difficult to explore the design space and find feasible solutions that have better objective values efficiently, causing optimization to stall.
\par
\begin{figure}[htbp]
  \begin{minipage}[b]{0.45\linewidth}
    \centering
    \includegraphics[scale=1.0] {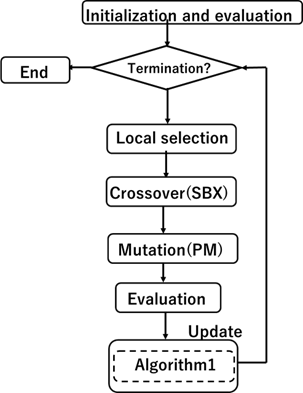}
    \caption{Flowchart of CMOEA/D. }
    \label{fig:flowCMOEAD}
  \end{minipage}
  \begin{minipage}[b]{0.45\linewidth}
    \centering
    \includegraphics[scale=1.0] {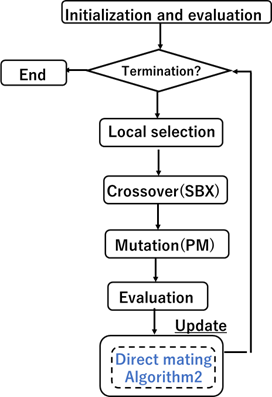}
    \caption{Flowchart of CMOEA/D-DMA. }
    \label{fig:flowCMOEAD-DMA}
  \end{minipage}
\end{figure}

To solve the problem of infeasible solutions not being utilized in optimization, DM, whose conceptual diagram is shown in Fig. 1, has been proposed. 
Figure \ref{fig:flowCMOEAD-DMA} shows the flow chart of CMOEA/D-DMA which was implemented DM to MOEA/D. 
Algorithm 2 shows the solution update. This is an individual selection method in which infeasible solutions with superior objective values are stored separately as archives $A_j$ and are used when generating offspring. This individual selection method incorporates design variables that can contribute to improving objective values by selecting infeasible solutions with excellent objective values. 
\par
A method that introduces this DM concept into CMOEA/D is called CMOEA/D-DMA\cite{miyakawa2018directed}. 
It is based on the conventional CMOEA/D method and separately archives solutions with superior objective values including infeasible solutions in archive $A_j (j = 1, 2,...,m)$. 
The archive size $\alpha$, which is the number of solutions to be kept as archive $A_j$, is a parameter set arbitrarily by the user. 
The constraint violation magnitude in DM is defined based on the domination in the constraint violation space to suppress bias in the solution search direction caused by differences in the scale of the constraint violation quantities. 
The dominance of solution ${\bm x}$ over ${\bm y}$ in terms of constraint violation quantities is denoted by ${\bm x} \succ cv{\bm y}$.

\begin{algorithm}
\caption{Solution update by the conventional CMOEA/D.}
\begin{algorithmic}
 \renewcommand{\algorithmicrequire}{\textbf{Input:}}
\renewcommand\algorithmicforall{\textbf{for each}}

\REQUIRE offspring ${\bf y}$, the focused index $i$
\FORALL {$j \in \mathcal{B}(i)$}
\IF{${\bf y}$ and ${\bf x}^j$ are feasible and $g({\bf y}|\lambda^j)$ is better than $g({\bf x}^j|\lambda^j)$}
  \STATE ${\bf x}^j \leftarrow \bf {\bf y}$ 
\ELSIF{${\bf y}$ is feasible and ${\bf x}^j$ is infeasible}
  \STATE ${\bf x}^j \leftarrow \bf {\bf y}$ 
\ELSIF{${\bf y}$ and ${\bf x}^j$ are feasible and $\Omega({\bf y})$ is less than $\Omega({\bf x}^j)$}
  \STATE ${\bf x}^j \leftarrow \bf {\bf y}$ 
\ENDIF
\ENDFOR
\end{algorithmic}
\end{algorithm}

\begin{algorithm}
\caption{Solution update by DMA}
\begin{algorithmic}
 \renewcommand{\algorithmicrequire}{\textbf{Input:}}
\renewcommand\algorithmicforall{\textbf{for each}}

\REQUIRE offspring ${\bf y}$, the focused index $i$
\FORALL {$j \in \mathcal{B}(i)$}
\IF{${\bf y}$ and ${\bf x}^j$ are feasible, and $g({\bf y}|\lambda^j)$ is better than $g({\bf x}^j|\lambda^j)$}
  \STATE ${\bf x}^j \leftarrow \bf {\bf y}$ 

\ELSIF{${\bf x}^j$ is feasible and ${\bf y}$ is infeasible, \\
and $g({\bf y}|\lambda^j)$ is better than $g({\bf x}^j|\lambda^j)$}
  \STATE $\mathcal{A}_j \leftarrow \mathcal{A}_j \cup {\bf y}$ 
  \IF{$|\mathcal{A}_j|>\alpha$}
  	\STATE Remove the worst solution from $\mathcal{A}_j$
  \ENDIF
    
\ELSIF{${\bf x}^j$ is infeasible and ${\bf y}$ is feasible}
  \STATE ${\bf x}^j \leftarrow \bf {\bf y}$ 
  
\ELSIF{${\bf x}^j$ and ${\bf y}$ are infeasible}
	\IF{${\bf y}$ dominates ${\bf x}^j$ on constraint violations}
  		\STATE ${\bf x}^j \leftarrow \bf {\bf y}$ 
  	\ELSIF{${\bf y}$ and ${\bf x}^j$ are non-dominated on constraint violations,\\
  		$g({\bf y}|\lambda^j)$ is better than $g({\bf x}^j|\lambda^j)$}
  		\STATE ${\bf x}^j \leftarrow \bf {\bf y}$ 
  	\ENDIF		
  
\ENDIF
\ENDFOR
\end{algorithmic}
\end{algorithm}


\section{PROPOSAL FOR IMPROVEMENT OF DIRECTIONAL MATING BY HYBRIDRIZATION OF LOCAL MATING}
While DM selects one parent from the dominated solutions for the first parent in the objective space at the archive update and achieves better exploration performance, the problem with DM is that it may not select appropriate solutions as a parent pair when there is no dominant solution. 
This is because DM requires solutions toward the optimal front from a firstly selected parent, which is more likely to occur towards the end of the exploration when many potential parents are located near the Pareto-optimal solutions. 
Additionally, in EA, the solution set at the end of the search generally depends on the distribution of the initial population, therefore the performance of DM depends on the initial population generated by random numbers.

\begin{figure}[htbp]
  \begin{minipage}[b]{0.45\linewidth}
    \centering
    \includegraphics[scale=1.0] {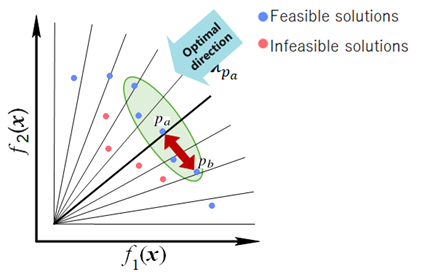}
    \caption{Conceptual diagram of local mating in CMOEA/D. Neighboorhood solutions is similary defined as the local selection procedure of MOEA/D, and only feasible solutions are selected. }
    \label{fig:DMA-NMselection}
  \end{minipage}
  \begin{minipage}[b]{0.45\linewidth}
    \centering
    \includegraphics[scale=1.0] {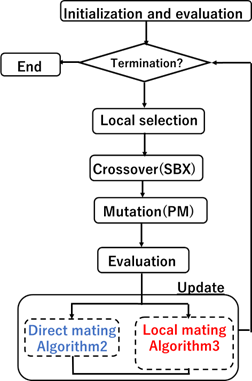}
    \caption{Flowchart of CMOEA/D-DMA-LM.}
    \label{fig:flowCMOEAD-DMA-LM}
  \end{minipage}
\end{figure}

To address this issue, we propose a hybridization of LM and DM. 
The proposed LM is another selection method in which the other parents are selected from feasible solutions that are in the neighborhood of the first selected parent. In the LM, the neighborhood solutions are defined similarly to the local selection procedure of MOEA/D, and only feasible solutions are selected in the focused weight vector $\mathcal{B}(i)$ of MOEA/D. 
Algorithm \ref{alg:NM} shows the update of the solutions by LM. The diversity of the solutions can be improved by LM in CMOEA/D-DMA. 
Figure \ref{fig:flowCMOEAD-DMA-LM} shows a conceptual diagram of LM.

In this study, we hybridize DM and LM into CMOEA/D and propose CMOEA/D-DMA-LM. 
The procedure for the proposed CMOEA/D-DMA-LM is shown in Figure \ref{fig:flowCMOEAD-DMA-LM}. 
This method has two selection methods, and a certain percentage of the solutions are selected by DM with a pre-defined probability r, while the others are selected by LM with a probability of $1-r$. 
In this study, $r$ is set to 0.5, which means that half of the solutions are selected by DM, and the other half are selected by neighborhood mating.
\begin{algorithm}
\caption{Solution update by LM.}
\label{alg:NM}
\begin{algorithmic}
 \renewcommand{\algorithmicrequire}{\textbf{Input:}}
\renewcommand\algorithmicforall{\textbf{for each}}

\REQUIRE offspring ${\bf y}$, the focused index $i$,$\mathcal{B}'(i)=\phi$
\FORALL {$j \in \mathcal{B}(i)$}
	\IF{${\bf x}^j$ is feasible,}
	\STATE $\mathcal{B}'(i) \leftarrow \mathcal{B}'(i) \in {\bf x}^j$
	\ENDIF
\ENDFOR

\FORALL {$j \in \mathcal{B}'(i)$}
\IF{${\bf y}$ and ${\bf x}^j$ are feasible and $g({\bf y}|\lambda^j)$ is better than $g({\bf x}^j|\lambda^j)$}
  \STATE ${\bf x}^j \leftarrow \bf {\bf y}$ 
\ELSIF{${\bf y}$ is feasible and ${\bf x}^j$ is infeasible}
  \STATE ${\bf x}^j \leftarrow \bf {\bf y}$ 
\ELSIF{${\bf y}$ and ${\bf x}^j$ are feasible and $\Omega({\bf y})$ is less than $\Omega({\bf x}^j)$}
  \STATE ${\bf x}^j \leftarrow \bf {\bf y}$ 
\ENDIF
\ENDFOR
\end{algorithmic}
\end{algorithm}

\section{NUMERICAL EXPERIMENT SETUP}
The mathematical problems OSY\cite{osyczka1995new},bi-objective mCDTLZ\cite{miyakawa2016controlling}, and TNK\cite{tanaka1995ga} were solved by CMOEA/D, CMOEA/D-DMA, and the proposed methods CMOEA/D-DMA-LM. 
In addition, the welded beam optimization problem (WB) \cite{deb2006reference} and the hybrid rocket engine design problem(HRE) \cite{kanazaki2017design} was solved for real-world industrial design applications. 
The average and standard deviation of the hypervolumes (HVs) for the non-dominated solutions with 20 attempts for OSY, mCDTLZ, TNK and WB with varying the random number for the initial individual generations in each generation, were used for the investigation. 
A smaller standard deviation indicated a smaller initial population dependency. 
For the calculation of HVs, the reference points were set as (-30.0,80.0) for OSY, (1.2,1.2) for TNK, (1.0,1.0) for mCDTLZ, (0.3,50.0) for WB, and (2000.0,0.0) for HRE.
The overview of the formulation and the design parameters for HREare discribed in the next subsection.
\par
This study compares CMOEA/D, CMOEA/D-DMA, and the proposed CMOEA/D-DMA-LM. Each method was performed for 1000 generations with 100 populations. 
For CMOEA/D-DMA-LM, the DM and LM rate r were set to 0.5. In each method, the simulated binary crossover (SBX) was used under crossover rates $Pc$ and the distance parameters of the crossover distribution $\eta_c$ are set to 0.9 and 20, respectively. 
The polynomial mutation was used when the mutation rate Pm was set to $1/n$, where $n$ is the total number of design variables.

\subsection{HRE: Multi-Disciplinary Optimization for the launch vehicle with hybrid rocket engine.}
HRE has a different combustion mechanism than conventional liquid and solid propulsion systems. 
Unlike liquid and solid propulsion systems, the oxidizer-to-fuel ratio (OF) for HRE must be determined after ignition. 
The geometry of the solid fuel and the supply control of the oxidizer can be optimized to improve the performance of the HRE. 

\par
In this study, we consider the optimization of the launch vehicle (LV) with HRE. 
Specifically, we aim to maximize the attained height ($H$) and minimize the maximum mass ($M_{\mathrm{tot}}$) of the LV. The LV is assumed to launch vertically. 
The design problem can be expressed as follows:

\begin{equation}
\left\{\begin{array}{l}
\mathrm{ Maximize } \quad H_{\rm max} \\
\mathrm{ Minimize } \quad M_{\mathrm{tot}}(0) \\
 \quad   \mathrm{Subject to} \\
 \quad   \quad  AR \leq 25.0[-] \\
 \quad   \quad  Q_{\rm max} \leq 100.0[\mathrm{kPa}] \\
 \quad   \quad Acc_{\max } \leq 5.0[\mathrm{G}] \\
\end{array}\right.
\end{equation}

\noindent
$AR$ represents the aspect ratio of the vehicle, which is constrained to accommodate arbitrary shape payloads. $Q_{\rm max}$ represents the maximum dynamic pressure, which is constrained to suppress aerodynamic vibrations, while $Acc_{\rm max}$ represents the maximum acceleration of the LV, which is constrained to ensure the safety of the payload.
This problem has six design variables such as LOX mass flow, fuel length, combustion time and chamber pressure.\cite{kanazaki2017design}

\begin{figure}[htbp]
    \centering
    \includegraphics[scale=0.9] {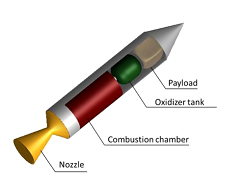}
      \caption{Schismatic illustration of a launch vehicle with HRE.}
    \label{fig:HR}
\end{figure}

\begin{table}[h]
 \caption{Design variables for HR powered LV}
 \label{table:SpeedOfLight}
 \centering
  \begin{tabular}{c|cccc}
    & Unit & Lower bound & Upper bound \\
       \hline
   dv1& LOX mass flow & ${\rm kg/s}$ &1.0 &30.0 \\
   dv2& Fuel length &  m &1.0 &10.0 \\ 
   dv3& Initial radius of fuel port &  mm & 10.0 &200.0 \\  
   dv4& Compustion time & s & 15.0 & 35.0 \\
   dv5& Chamber pressure & MPa & 3.0 & 4.00 \\ 
   dv6& Nozzle aperture ratio & - & 5.0 & 7.0 \\
   \hline
  \end{tabular}
\end{table}

\section{RESULTS and DISCUSSIONS}
\subsection{Investigation of the history of the non-dominated solutions by HVs}
\begin{figure}[htbp]
  \begin{minipage}[b]{0.3\linewidth}
    \centering
    \includegraphics[scale=0.55] {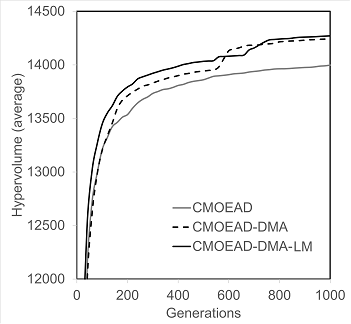}
  \end{minipage}
  \begin{minipage}[b]{0.3\linewidth}
    \centering
    \includegraphics[scale=0.55] {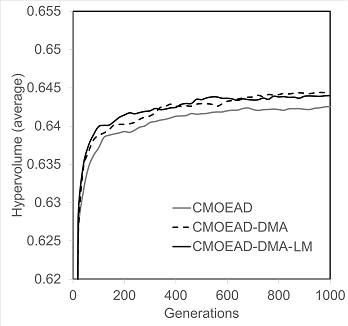}
  \end{minipage}
  \begin{minipage}[b]{0.3\linewidth}
    \centering
    \includegraphics[scale=0.55] {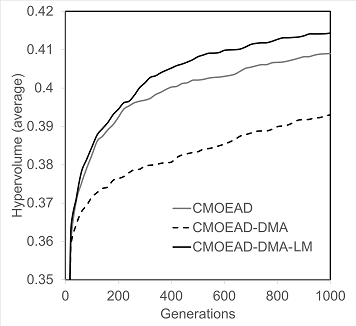}
  \end{minipage}
  \begin{minipage}[b]{0.3\linewidth}
    \centering
    \includegraphics[scale=0.55] {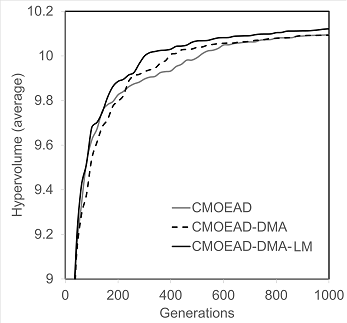}
  \end{minipage}
  \begin{minipage}[b]{0.3\linewidth}
    \centering
    \includegraphics[scale=0.55] {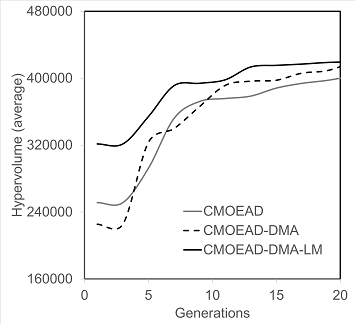}
  \end{minipage}
  \begin{minipage}[b]{0.3\linewidth}
    \centering    
  \end{minipage}
      \caption{The history of the averaged HVs among 20 attempts. (a) Results on OSY, (b) Results on TNK, (c) Results on mCDTLZ, and (d) real-world WB, and (e) real-world MDO problem, HRE(one attempt).}
    \label{fig:HVave}
\end{figure}

The histograms of the average HVs are shown in Fig. \ref{fig:HVave}. In each design problem, the proposed CMOEA/D-DMA-LM shows better convergence and a better final-generation HV value among the three methods.
\par
In the OSY results, as shown in Fig. \ref{fig:HVave}(a), CMOEA/D-DMA showed the same HV values in the final generation; however, the convergence did not improve.
OSY has six constraints, while the other three problems have two or three constraints and a non-smooth Pareto optimal set; thus, the proposed method can solve several constraint problems.
The HV histories of TNK are compared in Fig. \ref{fig:HVave}(b).
The proposed CMOEA/D-DMA-LM is superior to those of the other two methods in terms of HV values in the final generation and convergence.
TNK has a discontinuous Pareto front, and the maintenance of solution diversity is key to successful exploration. This result suggests that the proposed CMOEA/D-DMA-LM can maintain diversity by hybridizing the two selection methods, DM and LM. Owing to their high simulation cost, meta-heuristics cannot make several attempts at real-world design problems. Thus, this characteristic, the maintenance of diversity, is an advantage for solving real-world problems.
\par
The history of the HVs of mCDTLZ is shown in Fig. \ref{fig:HVave}(c). In this problem, the proposed CMOEA/D-DMA-LM obtains the best HV value in the last generation. The mCDTLZ has a Pareto front on the border between feasible and infeasible regions. This result suggests that the proposed CMOEA/D-DMA-LM can explore the edge of the feasible region, which is advantageous for real-world problems.
\par
The HV history of WB is shown in Fig. \ref{fig:HVave}(d). This problem simulates a real beam-welding design. The proposed CMOEA/D-DMA-LM shows better convergence and a better final-generation HV value among the three methods, as well as other mathematical test problems. This suggests that the proposed CMOEA/D-DMA-LM can effectively solve real-world design problems.
\par
The HV history of HRE is shown in Fig. \ref{fig:HVave}(e). 
This problem treats the multi-disciplinary design for the aerospace design problem. The proposed CMOEA/D-DMA-LM shows better convergence and a better final-generation HV value among the three methods. This suggests that the proposed CMOEA/D-DMA-LM can obtain better non-dominated solutions for such a complicated MDO problem. 
In other words, the proposed method is applicable to aerospace engineering design problems. For the HRE problem, we will carry out several attempts with changing the initial population to explore the solution space effectively.

\subsection{Initial population dependency by the standard deviation}
\begin{figure}[htbp]
  \begin{minipage}[b]{0.22\linewidth}
    \centering
    \includegraphics[scale=0.5]{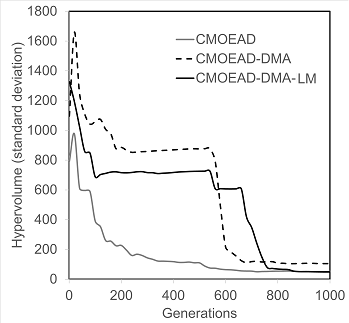}
  \end{minipage}
  \begin{minipage}[b]{0.22\linewidth}
    \centering
    \includegraphics[scale=0.5]{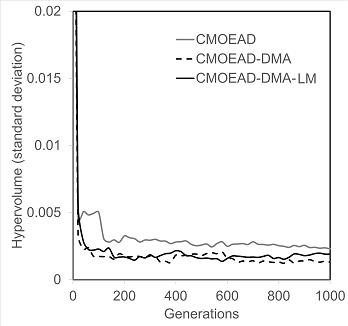}
  \end{minipage}
  \begin{minipage}[b]{0.22\linewidth}
    \centering
    \includegraphics[scale=0.5]{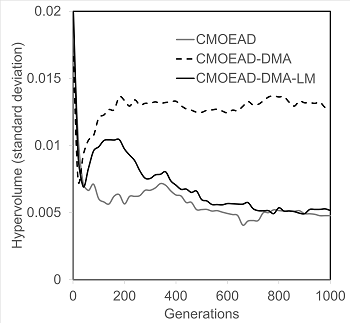}
  \end{minipage}
  \begin{minipage}[b]{0.22\linewidth}
    \centering
    \includegraphics[scale=0.5]{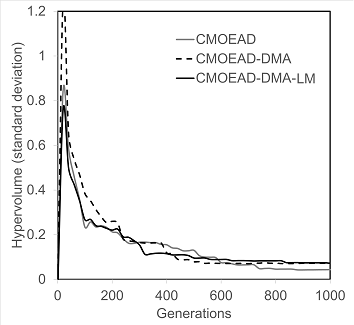}
  \end{minipage}
      \caption{The history of the std. of HVs among 20 attempts. (a) Results on OSY, (b) Results on TNK, (c) Results on mCDTLZ, and (d) real-world WB.}
    \label{fig:HVstd}
\end{figure}

The comparison of the standard deviation history of HVs with 20 attempts is shown in Fig. \ref{fig:HVstd}. The proposed MOEA/D-DMA-LM showed the smallest std values compared to any other methodology from the early generation for OSY and TNK problems. As described earlier, OSY has a non-smooth Pareto optimal set with many constraints, and TNK has a discontinuous Pareto optimal set. This suggests that the proposed MOEA/D-DMA-LM can minimize the initial population dependency of the HVs, even when the problem has a non-smooth Pareto optimal set.
In the result of mCDTLZ, as shown in Fig. \ref{fig:HVstd}(c), the proposed CMOEA/D-DMA-LM could minimize the HV’s std. approximately after 20 generations compared to CMOEA/D-DMA. mCDTLZ has an infeasible Pareto optimal set close to the infeasible region. This result suggests that the proposed CMOEA/D-DMA-LM can maintain the diversity of solutions. Moreover, CMOEA/D-DMA could not maintain diversity and showed a higher std. This result suggests that the DM has a disadvantage in maintaining diversity, and the LM has an advantage because it explores the infeasible region.
As discussed in the previous subsection, the average HV of the proposed CMOEA/D-DMA-LM was the highest. Thus, the proposed hybrid method can solve real-world problems well when the number of attempts with changing initial populations is limited due to the high computational cost of evaluating the fitness function.

\section{Conclutinos}
In this study, we propose an improved constraint-handling method called direct mating-based CMOEA/D-DMA-LM, which hybridizes direct mating with neighborhood mating to generate diverse solutions. The method selects half of the population by direct mating and the other half by neighborhood mating, aiming to balance exploration and exploitation. We compare the proposed method with existing CMOEA/D and CMOEA/D-DMA methods on three mathematical test problems and one industrial problem. The performance of each method is evaluated using the average hypervolume over 20 runs and the standard deviation of the hypervolume. Our results show that the proposed CMOEA/D-DMA-LM outperforms the other methods in terms of hypervolume values and convergence rate. Additionally, the proposed method exhibits better performance in terms of standard deviation, initial population dependency, and diversity along the generation in two test functions, OSY and TNK. However, in one problem (mCDTLZ), the existing CMOEA/D obtains a smaller standard deviation in the earlier generation. For the industrial problem (WB), each method has a similar standard deviation history. Therefore, the proposed CMOEA/D-DMA-LM method can obtain higher average hypervolumes while maintaining diversity. 
\par
To further improve our method, we will investigate the optimal ratio of direct mating to neighborhood mating and suitable crossover methods. For example, we may consider using differential evolution for neighborhood mating, which can maintain high diversity by improving solution diversity.


\bibliographystyle{unsrt}  
\bibliography{EA-EGO-ref_r220405}

\end{document}